\pdfoutput=1

\documentclass[11pt]{article}

\usepackage[]{acl}

\usepackage{times}
\usepackage{latexsym}

\usepackage[T1]{fontenc}

\usepackage[utf8]{inputenc}

\usepackage{microtype}
\usepackage{booktabs}
\usepackage{graphicx}
\usepackage{multirow}
\usepackage{amssymb}

\title{Overview of the BioLaySumm 2024 Shared Task on the Lay Summarization of Biomedical Research Articles}

\author{Tomas Goldsack$^{1}$, \textbf{Carolina Scarton}$^{1}$, \textbf{Matthew Shardlow}$^{3}$, \textbf{Chenghua Lin}$^{1,2}$ \\
        $^{1}$University of Sheffield, $^{2}$University of Manchester,
        $^{3}$Manchester Metropolitan University \\
        \texttt{\{tgoldsack1, c.lin, c.scarton\}@sheffield.ac.uk}\\
        \texttt{m.shardlow@mmu.ac.uk}}

\begin{document}

\maketitle
\begin{abstract}

This paper presents the setup and results of the second edition of the BioLaySumm shared task on the Lay Summarisation of Biomedical Research Articles, hosted at the BioNLP Workshop at ACL 2024.  
In this task edition, we aim to build on the first edition's success by further increasing research interest in this important task and encouraging participants to explore novel approaches that will help advance the state-of-the-art. 
Encouragingly, we found research interest in the task to be high, with this edition of the task attracting a total of 53 participating teams, a significant increase in engagement from the previous edition. Overall, our results show that a broad range of innovative approaches were adopted by task participants, with a predictable shift towards the use of Large Language Models (LLMs).

\end{abstract}

\section{Introduction}
Lay Summarisation describes the task of generating a summary of a technical or specialist text that is suitable for a non-expert audience. To achieve this goal, a good lay summary will typically focus on explaining the relevant background information alongside the significance and findings of an article, while avoiding extensive use of jargon or technical language. As such, lay summaries offer significant benefits in broadening access to technical articles that are of interest to a broad range of audiences. 

Biomedical research publications, containing the latest research on prominent health-related topics, represent a perfect example of such texts. Not only are the contents of these articles relevant to other researchers working in the same domain, but often they can also be of interest to researchers in related domains, medical practitioners, and even members of the public (e.g., those affected by an illness/disease being studied). In this scenario, the lay summary is an essential tool in allowing these secondary audiences, who don't possess the expertise required to interpret the full article, to understand its key findings and relevance to their information needs.

Although Lay Summaries are required or encouraged by some biomedical publications, they are not universal, leaving a significant amount of inaccessible to lay audiences. Furthermore, the burden of writing these summaries is often placed upon the article authors, who are not always adept at effectively communicating their work to a non-technical audience.
As such, automatic lay summary generation has significant potential to help both authors and non-expert readers by improving the dissemination of important research.



The BioLaySumm shared task\footnote{\url{https://biolaysumm.org}} 
focuses on improving the automatic lay summarization of biomedical research. 
Through this shared task, we aim to encourage the development of novel approaches and increase interest in the research of automatic techniques for disseminating scientific research to broad audiences.  
In this paper, we present the results of the second edition of the BioLaySumm shared task, hosted by the BioNLP Workshop at ACL 2024.

In what remains of the paper, we address the task formulation (\S\ref{sec:tasks}), datasets (\S\ref{sec:data}), and evaluation procedure (\S\ref{sec:eval}), before providing a description of overall results and notable insights (\S\ref{sec:results}), and finall the participating systems (\S\ref{sec:submission}).

\section{Task Description} \label{sec:tasks}


As part of the task, participants must develop systems capable of generating a lay summary of biomedical research, given the article's text as input. Our competition was hosted using the CodaBench platform \citep{XU2022100543}.

As with the previous edition of the task, two separate datasets, \textbf{PLOS} and \textbf{eLife} are used. Participants were provided with both training and validation sets, complete with reference lay summaries, alongside a blind test set. Final system performance is determined by the performance of participants' systems on the blind test set, which could be obtained by submitting their predicted lay summaries to our CodaBench competition, where they were automatically evaluated.   
More information regarding our datasets and evaluation protocol is provided in subsequent sections (\S\ref{sec:data} and \S\ref{sec:eval}, respectively).

We allowed submissions to be generated from either two separate summarisation models (i.e., one trained on each dataset) or a single unified model (i.e., trained on both datasets). Participants were required to indicate which approach was taken for each submission, in addition to whether or not they made use of additional training data (i.e., data not provided specifically for the task).
Participants were also allowed to compete as part of teams, where each team was permitted to make a maximum of 15 test set submissions to the CodaBench platform.\footnote{A significant increase on the limit of 3 submissions imposed in the previous edition of the task.} However, teams were required to select only one of their submissions to appear on the final task leaderboard. 

Because of its strong performance in the previous edition of the task, we also choose to keep the same baseline system. Specifically, this baseline system consists of two separate BART-base models \citep{lewis-etal-2020-bart}, trained independently on the PLOS and eLife datasets. 

\section{Datasets} \label{sec:data}

The datasets used for the task are based on the previous works of \citet{goldsack-etal-2022-making} and \citet{luo-etal-2022-readability}, and are derived from two different biomedical publications: \textbf{Public Library of Science (PLOS)} and \textbf{eLife}. Each dataset consists of biomedical research articles paired with expert-written lay summaries.

As described in \citet{goldsack-etal-2022-making}, the lay summaries of each dataset also exhibit numerous notable differences in their characteristics, with eLife's lay summaries being longer, more abstractive, and more readable than those of PLOS.

Furthermore, for PLOS, lay summaries are author-written, and articles are derived from 5 peer-reviewed journals covering Biology, Computational Biology, Genetics, Pathogens, and Neglected Tropical Diseases. For eLife, lay summaries are written by expert editors (in correspondence with authors), and articles are derived from the peer-reviewed eLife journal, covering all areas of the life sciences and medicine. For a more detailed analysis of dataset content, readers can refer to \citet{goldsack-etal-2022-making}.

\begin{table}[]
    \centering
    \begin{tabular}{lcccc}
         \hline
         \textbf{Dataset} & \textbf{Subtask} & \textbf{\# Train} & \textbf{\# Val} & \textbf{\# Test}   \\ \hline
         eLife & 1 & 4,346 & 241 & 142 \\
         PLOS & 1, 2 & 24,773 & 1,376 & 142* \\ \hline
    \end{tabular}
    \caption{Data split sizes for each dataset. * denotes that this split is different for each subtask.}
    \label{tab:data}
\end{table}

Table \ref{tab:data} summarises the data split information for both datasets. Note that the training and validation sets used for both datasets are equal to those published in \citet{goldsack-etal-2022-making}. 

As with the previous task edition, we collect new test splits for both PLOS and eLife data using more recently published articles from each respective journal.
This test data consists of 142 PLOS articles and 142 eLife articles.

In utilizing these datasets for our task, we hope to enable the training of abstractive summarisation models that are capable of generating lay summaries for unseen articles covering a wide range of biomedical topics, enabling the significance of new, important publications to be effectively communicated to non-expert audiences.

\section{Evaluation}\label{sec:eval}



For both subtasks, we evaluate summary quality according to three criteria - \textit{Relevance}, \textit{Readability}, and \textit{Factuality} - where each criterion is composed of one or more automatic metrics:

\begin{itemize}
  \item \textit{Relevance}: ROUGE-1, 2, and L \citep{lin-2004-rouge} and BERTScore \citep{ZhangKWWA20}.
  \item \textit{Readability}: Flesch-Kincaid Grade Level (FKGL), Dale-Chall Readability Score (DCRS), *Coleman-Laiu Index (CLI), and *LENS \citep{maddela-etal-2023-lens}.
  \item \textit{Factuality}: *AlignScore \citep{zha-etal-2023-alignscore} and *SummaC \citep{zha-etal-2023-alignscore}
\end{itemize}

Where ``*" indicates that the metric is newly introduced for this year's edition of the task.
Specifically, the CLI and LENS metrics are introduced in order to enhance our evaluation of summary readability. Alternatively, AlignScore and SummaC are introduced to replace the fine-tuned BARTScore model used to assess factuality in the previous task edition, with the reason for this being that BARTScore was found to exhibit bias toward BART-based approaches.

The scores calculated for each metric are the average of those calculated independently for the generated lay summaries of PLOS and eLife. The aim is to maximize the scores for all metrics except for FKGL, DCRS, and CLI the  Readability metrics. For these metrics, the aim is to minimize scores, as lower scores are indicative of greater readability.\footnote{For these metrics, the scores are estimates of the US Grade level of education required to comprehend a given text.}

Following the submission deadline for each subtask, an overall ranking is calculated based on the average performance of submissions across all criteria. Specifically, we first apply min-max normalization to the scores of each metric (thus establishing a common value range), before averaging across metrics within each criterion to obtain criterion-level scores.\footnote{This represents a minor change from the averaging protocol of the previous task edition, in which we first calculated rankings for each criterion, before summing these rankings to compute an overall rank.} Note that, for metrics that we minimize (i.e., FKGL, DCRS, and CLI) we calculate 1 minus the mix-max normalized value. Finally, criterion-level scores are then averaged to obtain an overall score, by which submissions are then ranked.

\begin{table*}[]
    \centering
    \resizebox{\textwidth}{!}{
    \begin{tabular}{rlccccccccccccccc}
        \hline \multirow{2}{*}{\textbf{$\star$}} & \multirow{2}{*}{\textbf{Team}}  & \multirow{2}{*}{\textbf{\#}} & \multirow{2}{*}{\textbf{+}} & \multicolumn{4}{c}{\textbf{Relevance}} && \multicolumn{4}{c}{\textbf{Readability}} &&  \multicolumn{2}{c}{\textbf{Factuality}}  \\ \cline{5-8} \cline{10-13} \cline{15-16}
        &&&& \textbf{R-1} & \textbf{R-2} & \textbf{R-L} & \textbf{BertS} && \textbf{FKGL} & \textbf{DCRS} & \textbf{CLI} & \textbf{LENS} && \textbf{AlignS} & \textbf{SummaC} \\
           \hline
         1  & \textbf{UIUC\_BioNLP} & 2 & $\times$ & 48.55 & \textbf{15.69} & \textbf{45.50} & \textbf{86.77} && 11.75 & 9.34 & 13.36 & 52.85 && 80.04 & 73.38  \\
         2  & \textbf{Ctyun AI} & 2 & $\times$  & 47.96 & 15.46 & 44.94 & 86.66 && 12.44 & 9.67 & 14.15 & 51.09 && 82.72 & 74.80 \\
         3  & \textbf{Saama Technologies} & 2 & $\times$ & 47.85 & 15.45 & 44.97 & 86.70 && 11.36 & 9.10 & 13.15 & 51.90 && 77.83 & 72.68 \\
         4  & \textbf{WisPerMed} & 2 & $\times$ & 47.12 & 15.18 & 44.28 & 86.53 && 11.07 & 8.86 & 12.87 & 51.03 && 78.18 & 72.16 \\
         5  & \textbf{cylaun} & 1 & $\times$ & 47.39 & 14.55 & 44.45 & 85.61 && \textbf{10.46} & 9.33 & 12.64 & 41.69 && 75.26 & 78.44  \\
         6 & \underline{\textbf{BART Baseline}} & 2 & $\times$ & 46.96 & 13.95 & 43.58 & 86.23 && 12.04 & 10.15 & 13.49 & 48.10 && 77.88 & 70.26  \\
         7 & \textbf{AUTH} & 1 & $\checkmark$ & 48.23 & 14.57 & 44.77 & 85.76 && 12.44 & 10.04 & 13.50 & 66.11 && 74.18 & 66.40 \\
         8 & \textbf{maverick} & 1 & $\times$ &  42.77 & 12.97 & 39.42 & 85.01 && 15.04 & 10.65 & 16.61 & 52.30 && 91.22 & 83.85 \\
         9 & \textbf{Empress} & 1 & $\times$ & 43.96 & 12.29 & 41.36 & 84.89 && 10.66 & 9.06 & 12.89 & 59.73 && 73.47 & 68.02 \\
         10 & \textbf{eulerian} & 1 & $\times$ & 40.35 & 11.66 & 37.10 & 84.51 && 14.80 & 10.76 & 16.53 & 48.46 && 91.73 & 85.38 \\
         11 & \textbf{BioLay\_AK\_SS} & 2 & $\times$ & 43.98 & 12.15 & 40.39 & 84.71 && 14.20 & 11.12 & 15.12 & 49.57 && 85.03 & 78.60 \\
         12 & \textbf{HULAT-UC3M} & 2 & $\times$ & \textbf{48.72} & 14.65 & 45.20 & 86.22 && 12.71 & 10.43 & 14.08 & 49.34 && 66.69 & 67.03 \\
         13 & \textbf{Atif\_Tanish} & 1 & $\times$ & 43.82 & 11.96 & 41.01 & 84.84 && 10.61 & 9.12 & 12.86 & 60.14 && 72.92 & 67.12\\
         14 & \textbf{qwerty} & 1 & $\times$ & 37.26 & 10.45 & 34.48 & 83.54 && 13.36 & 9.18 & 14.60 & 42.16 && 89.89 & 83.23 \\
         15 & \textbf{Deakin} & 2 & $\times$ & 48.22 & 14.20 & 44.41 & 85.83 && 14.46 & 10.76 & 15.48 & 63.91 && 74.57 & 61.80 \\
         16 & \textbf{MDSCL} &  2 & $\times$ & 42.56 & 13.01 & 39.35 & 85.20 && 14.01 & 10.78 & 15.92 & 63.05 && 81.50 & 71.54 \\
         17 & \textbf{MDS-CL}  & 2 & $\times$ & 42.13 & 12.90 & 38.93 & 85.14 && 14.13 & 10.82 & 15.96 & 61.71 && 81.98 & 73.14 \\
         18 & \textbf{elirf} & 2 & $\checkmark$ & 48.15 & 13.66 & 43.09 & 85.95 && 13.61 & 10.86 & 14.66 & 48.02 && 78.21 & 60.66 \\
         19 & \textbf{RAG-RLRC-LaySum} & 2 & $\times$ & 46.24 & 13.04 & 42.37 & 85.29 && 12.68 & 10.43 & 14.41 & 59.26 && 71.28 & 66.29 \\
         20 & \textbf{naive\_bhais} & 2 & $\times$ & 43.42 & 12.60 & 39.91 & 85.72 && 12.89 & 10.94 & 14.32 & 37.86 && 81.34 & 67.81 \\
         21 & \textbf{MDS-CL} & 1 & $\times$ & 42.31 & 11.05 & 39.22 & 85.62 && 11.93 & 9.23 & 13.25 & 74.67 && 71.52 & 56.55 \\
         22 & \textbf{MDS-CL} & 1 & $\times$ & 43.43 & 11.98 & 40.13 & 85.55 && 12.39 & 9.76 & 14.28 & 76.80 && 72.18 & 54.41 \\
         23 & \textbf{DhruvShlo} & 1 & $\times$ & 42.15 & 11.05 & 39.40 & 84.42 && 11.76 & 9.08 & 13.02 & 49.17 && 71.25 & 63.98   \\
         24 & \textbf{naman\_tejas} & 1 & $\times$ & 39.54 & 11.06 & 36.73 & 84.25 && 12.29 & 9.20 & 13.58 & 50.44 && 75.68 & 68.10  \\
         25 & \textbf{SINAI} & 2 & $\times$ & 42.05 & 12.49 & 38.53 & 85.83 && 12.23 & 9.86 & 13.81  & 76.95 && 71.17 & 53.98 \\
         26 & \textbf{XYZ} & 2 & $\times$ & 41.04 & 9.93 & 38.01 & 85.50 && 11.02 & 9.37 & 13.00 & \textbf{81.21} && 70.18 & 54.63 \\
         27 & \textbf{gpsigh} & 2 & $\times$ & 33.60 & 9.18 & 30.97 & 82.97 && 15.69 & 9.30 & 15.17 & 42.06 && 91.28 & 82.11 \\
         28 & \textbf{YXZ} & 2 & $\times$ & 42.25 & 10.91 & 39.20 & 84.99 && 11.18 & 8.57 & 12.44 & 71.57 && 64.89 & 53.49 \\
         29 & \textbf{sanika} & 2 & $\times$ & 42.90 & 11.16 & 38.06 & 83.33 && 17.93 & 12.40 & 17.37 & 11.37 && 85.16 & \textbf{90.28} \\
         30 & \textbf{Bossy Beaver} & 1 & $\times$ & 41.32 & 11.45 & 37.98 & 84.73 && 13.99 & 10.41 & 15.74 & 65.49 && 78.08 & 60.65  \\
         31 & \textbf{Dayal K-Laksh G} & 1 & $\times$ & 33.93 & 9.49 & 30.55 & 84.98 && 14.39 & 12.15 & 16.24 & 32.33 && 93.07 & 80.71 \\
         32 & \textbf{MKGS} &  1 & $\times$ & 37.75 & 9.72 & 34.67 & 83.33 && 15.79 & 11.92 & 17.50 & 22.07 && \textbf{93.08} & 83.52 \\
         33 & \textbf{Shallow-Learning} &  1 & $\times$ & 42.22 & 11.33 & 39.54 & 83.89 && 10.56 & 9.04 & 12.42 & 53.68 && 57.28 & 61.17 \\
         34 & \textbf{NLPSucks} & 2 & $\times$ & 34.91 & 8.32 & 33.32 & 82.62 && 10.68 & \textbf{6.76} & 12.08 & 37.86 && 74.36 & 64.22 \\
         35 & \textbf{CookieMonster} & 2 & $\times$ & 43.06 & 10.33 & 39.83 & 84.57 && 12.04 & 9.37 & 13.18 & 49.53 && 63.63 & 59.19 \\
         36 & \textbf{NoblesseUranium} & 1 & $\times$ & 39.16 & 10.34 & 35.87 & 84.65 && 14.21 & 10.44 & 15.45 & 51.99 && 75.74 & 67.32 \\
         37 & \textbf{roon} & 2 & $\times$ & 44.16 & 11.24 & 41.44 & 84.74 && 11.78 & 8.86 & 12.38 & 71.26 && 52.73 & 50.50\\
         38 & \textbf{jimmyapples} & 2 & $\times$ & 43.36 & 10.84 & 40.35 & 84.82 && 11.44 & 9.03 & 12.10 & 71.48 && 56.54 & 49.00  \\
         39 & \textbf{Shivam} & 2 & $\times$ & 33.85 & 8.91 & 30.76 & 83.56 && 12.90 & 11.89 & 15.14 & 15.66 && 91.64 & 80.94 \\
         40 & \textbf{HGP\_NLP} & 2 & $\times$ & 29.69 & 8.60 & 26.95 & 83.74 && 11.20 & 9.91 & 12.79 & 44.22 && 79.45 & 74.11 \\
         41 & \textbf{Cornell-BioLay} &  1 & $\times$ & 39.50 & 7.92 & 35.99 & 84.50 && 10.97 & 9.56 & 12.66 & 72.53 && 60.10 & 51.76 \\
         42 & \textbf{xpc} & 2 & $\times$ & 44.59 & 11.80 & 40.36 & 84.84 && 13.45 & 10.33 & 15.72 & 67.72 && 56.87 & 48.78 \\
         43 & \textbf{Hemlo} & 1 & $\times$ & 30.04 & 6.88 & 27.86 & 81.10 && 16.49 & 7.58 & 15.74 & 21.91 && 89.50 & 74.41   \\
         44 & \textbf{anjaneya} & 2 & $\times$ & 28.87 & 8.26 & 26.00 & 83.66 && 13.70 & 11.45 & 15.46 & 37.03 && 74.71 & 78.66 \\
         45 & \textbf{Runtime\_Terror} & 1 & $\times$ & 40.18 & 10.14 & 37.44 & 83.69 && 13.98 & 8.41 & 13.13 & 49.60 && 47.51 & 50.37 \\
         46 & \textbf{Abhi\_Sidd} &  1 & $\times$ & 35.33 & 9.15 & 31.79 & 83.09 && 17.29 & 12.42 & 14.47 & 17.41 && 79.09 & 62.95 \\
         47 & \textbf{cbdch} & 1 & $\times$ & 36.69 & 9.29 & 32.98 & 85.09 && 14.43 & 11.18 & 15.82 & 74.13 && 63.34 & 44.61 \\
         48 & \textbf{aLoneLM} &  1 & $\times$ & 35.67 & 6.74 & 33.29 & 82.33 && 11.07 & 8.52 & \textbf{11.04} & 42.82 && 48.93 & 52.62 \\
         49 & \textbf{hohoho} & 2 & $\times$ & 33.46 & 5.48 & 31.32 & 81.93 && 11.21 & 8.80 & 12.04 & 53.01 && 44.68 & 50.98 \\
         50 & \textbf{huizige} & 1 & $\times$ & 37.16 & 8.82 & 33.59 & 83.06 && 15.40 & 11.39 & 16.78 & 47.53 && 60.13 & 49.36 \\
         51 & \textbf{SSS} & 1 & $\times$ & 25.64 & 6.21 & 23.18 & 82.81 && 13.71 & 12.45 & 16.61 & 43.82 && 72.72 & 61.55 \\
         52 & \textbf{H2P} &  1 & $\times$ & 25.93 & 4.03 & 23.73 & 81.70 && 16.53 & 11.94 & 18.98 & 56.77 && 56.03 & 47.04 \\
         53 & \textbf{KnowLab} & 1 & $\checkmark$ & 32.16 & 7.34 & 28.31 & 80.63 && 36.29 & 11.55 & 11.28 & 1.32 && 42.41 & 54.74 \\ \hline
    \end{tabular}
    }
    \caption{Task leaderboard - all metrics. The $\star$ column denotes the submission rank, the \textbf{\#} column the number of models used - 1 (unified) or 2 (one for each dataset), and the \textbf{+} column the use of additional training data. \textbf{R} = ROUGE F1, \textbf{BertS} = BertScore, \textbf{FKGL} = Flesch-Kincaid Grade Level, \textbf{DCRS} = Dale-Chall Readability Score, \textbf{CLI} = Coleman-Liau Index, \textbf{AlignS} = AlignScore.
    }
    \label{tab:st1_results}
\end{table*}

\begin{table*}[]
    \centering
    \resizebox{\textwidth}{!}{
    \begin{tabular}{rlccccccccccccc}
        \hline \multirow{2}{*}{\textbf{Rank}} & \multirow{2}{*}{\textbf{Team}} & \multicolumn{4}{c}{\textbf{Relevance}} && \multicolumn{4}{c}{\textbf{Readability}} &&  \multicolumn{2}{c}{\textbf{Factuality}}  \\ \cline{3-6} \cline{8-11} \cline{13-14}
        && \textbf{R-1} & \textbf{R-2} & \textbf{R-L} & \textbf{BertS} && \textbf{FKGL} & \textbf{DCRS} & \textbf{CLI} & \textbf{LENS} && \textbf{AlignS} & \textbf{SummaC} \\
           \hline
            1	&	\textbf{UIUC\_BioNLP}	&	0.993	&	\textbf{1.000}	&	\textbf{1.000}	&	\textbf{1.000}	&&	0.950	&	0.547	&	0.708	&	0.645	&&	0.743	&	0.630	\\
            2	&	\textbf{Ctyun AI}  \	&	0.967	&	0.980	&	0.975	&	0.982	&&	0.923	&	0.488	&	0.609	&	0.623	&&	0.796	&	0.661	\\
            3	&	\textbf{Saama Technologies} 	&	0.962	&	0.979	&	0.976	&	0.989	&&	0.965	&	0.589	&	0.734	&	0.633	&&	0.699	&	0.615	\\
            4	&	\textbf{WisPerMed} 	&	0.931	&	0.956	&	0.945	&	0.962	&&	0.976	&	0.631	&	0.770	&	0.622	&&	0.706	&	0.603	\\
            5	&	\textbf{cylaun} 	&	0.943	&	0.902	&	0.953	&	0.811	&&	\textbf{1.000}	&	0.548	&	0.800	&	0.505	&&	0.648	&	0.741	\\
            6	&	\underline{\textbf{BART Baseline}}	&	0.924	&	0.851	&	0.914	&	0.913	&&	0.939	&	0.405	&	0.693	&	0.586	&&	0.700	&	0.561	\\
            7	&	\textbf{AUTH}	&	0.979	&	0.904	&	0.967	&	0.836	&&	0.923	&	0.424	&	0.691	&	0.811	&&	0.627	&	0.477	\\
            8	&	\textbf{maverick}	&	0.742	&	0.766	&	0.728	&	0.713	&&	0.822	&	0.316	&	0.298	&	0.638	&&	0.963	&	0.859	\\
            9	&	\textbf{Empress}	&	0.794	&	0.708	&	0.815	&	0.695	&&	0.992	&	0.596	&	0.768	&	0.731	&&	0.613	&	0.513	\\
            10	&	\textbf{eulerian}	&	0.637	&	0.655	&	0.624	&	0.632	&&	0.832	&	0.298	&	0.308	&	0.590	&&	0.973	&	0.893	\\
            11	&	\textbf{BioLay\_AK\_SS}	&	0.795	&	0.697	&	0.771	&	0.664	&&	0.855	&	0.233	&	0.486	&	0.604	&&	0.841	&	0.744	\\
            12	&	\textbf{HULAT-UC3M}	&	\textbf{1.000}	&	0.911	&	0.987	&	0.912	&&	0.913	&	0.355	&	0.618	&	0.601	&&	0.479	&	0.491	\\
            13	&	 \textbf{Atif\_Tanish}	&	0.788	&	0.680	&	0.799	&	0.686	&&	0.994	&	0.585	&	0.771	&	0.736	&&	0.602	&	0.493	\\
            14	&	\textbf{qwerty}	&	0.503	&	0.551	&	0.506	&	0.475	&&	0.888	&	0.574	&	0.552	&	0.511	&&	0.937	&	0.845	\\
            15	&	\textbf{Deakin}	&	0.978	&	0.872	&	0.951	&	0.848	&&	0.845	&	0.298	&	0.441	&	0.784	&&	0.635	&	0.376	\\
            16	&	\textbf{MDSCL}	&	0.733	&	0.770	&	0.725	&	0.745	&&	0.862	&	0.294	&	0.385	&	0.773	&&	0.771	&	0.590	\\
            17	&	\textbf{MDS-CL}	&	0.714	&	0.761	&	0.706	&	0.734	&&	0.858	&	0.286	&	0.381	&	0.756	&&	0.781	&	0.625	\\
            18	&	\textbf{elirf}	&	0.976	&	0.826	&	0.892	&	0.867	&&	0.878	&	0.280	&	0.544	&	0.585	&&	0.707	&	0.351	\\
            19	&	\textbf{RAG-RLRC-LaySum}	&	0.892	&	0.772	&	0.860	&	0.759	&&	0.914	&	0.355	&	0.576	&	0.725	&&	0.570	&	0.475	\\
            20	&	\textbf{naive\_bhais}	&	0.790	&	0.749	&	0.773	&	0.850	&&	0.912	&	0.270	&	0.578	&	0.498	&&	0.760	&	0.493	\\
            21	&	\textbf{MDS-CL}	&	0.722	&	0.601	&	0.719	&	0.813	&&	0.943	&	0.567	&	0.722	&	0.918	&&	0.575	&	0.261	\\
            22	&	\textbf{MDS-CL} 	&	0.771	&	0.682	&	0.759	&	0.802	&&	0.925	&	0.473	&	0.592	&	0.945	&&	0.588	&	0.214	\\
            23	&	\textbf{DhruvShlo}	&	0.716	&	0.602	&	0.727	&	0.618	&&	0.950	&	0.593	&	0.751	&	0.599	&&	0.569	&	0.424	\\
            24	&	\textbf{naman\_tejas}	&	0.602	&	0.603	&	0.607	&	0.589	&&	0.929	&	0.571	&	0.681	&	0.615	&&	0.657	&	0.514	\\
            25	&	\textbf{SINAI}	&	0.711	&	0.726	&	0.688	&	0.848	&&	0.932	&	0.455	&	0.651	&	0.947	&&	0.568	&	0.205	\\
            26	&	\textbf{XYZ}	&	0.667	&	0.506	&	0.665	&	0.793	&&	0.978	&	0.542	&	0.754	&	\textbf{1.000}	&&	0.548	&	0.219	\\
            27	&	\textbf{gpsigh}	&	0.345	&	0.442	&	0.349	&	0.381	&&	0.798	&	0.553	&	0.481	&	0.510	&&	0.965	&	0.821	\\
            28	&	\textbf{YXZ}	&	0.720	&	0.590	&	0.718	&	0.711	&&	0.972	&	0.682	&	0.825	&	0.879	&&	0.444	&	0.194	\\
            29	&	\textbf{sanika}	&	0.748	&	0.612	&	0.667	&	0.440	&&	0.711	&	0.008	&	0.203	&	0.126	&&	0.844	&	\textbf{1.000}	\\
            30	&	\textbf{Bossy Beaver} 	&	0.679	&	0.636	&	0.663	&	0.668	&&	0.863	&	0.359	&	0.409	&	0.803	&&	0.704	&	0.351	\\
            31	&	\textbf{Dayal K-Laksh G}	&	0.359	&	0.468	&	0.330	&	0.708	&&	0.848	&	0.053	&	0.346	&	0.388	&&	\textbf{1.000}	&	0.790	\\
            32	&	\textbf{MKGS}	&	0.525	&	0.488	&	0.515	&	0.440	&&	0.794	&	0.093	&	0.187	&	0.260	&&	\textbf{1.000}	&	0.852	\\
            33	&	\textbf{Shallow-Learning}	&	0.718	&	0.626	&	0.733	&	0.531	&&	0.996	&	0.600	&	0.826	&	0.655	&&	0.293	&	0.362	\\
            34	&	\textbf{NLPSucks} 	&	0.402	&	0.368	&	0.454	&	0.324	&&	0.991	&	\textbf{1.000}	&	0.870	&	0.457	&&	0.631	&	0.429	\\
            35	&	\textbf{CookieMonster}	&	0.755	&	0.540	&	0.746	&	0.643	&&	0.939	&	0.541	&	0.731	&	0.604	&&	0.419	&	0.319	\\
            36	&	\textbf{NoblesseUranium}	&	0.586	&	0.541	&	0.568	&	0.656	&&	0.855	&	0.353	&	0.445	&	0.634	&&	0.658	&	0.497	\\
            37	&	\textbf{roon}	&	0.802	&	0.618	&	0.818	&	0.670	&&	0.949	&	0.632	&	0.832	&	0.875	&&	0.204	&	0.129	\\
            38	&	\textbf{jimmyapples} 	&	0.768	&	0.584	&	0.769	&	0.683	&&	0.962	&	0.601	&	0.867	&	0.878	&&	0.279	&	0.096	\\
            39	&	\textbf{Shivam} 	&	0.356	&	0.418	&	0.340	&	0.477	&&	0.905	&	0.098	&	0.484	&	0.179	&&	0.972	&	0.795	\\
            40	&	\textbf{HGP\_NLP}	&	0.175	&	0.392	&	0.169	&	0.507	&&	0.971	&	0.446	&	0.781	&	0.537	&&	0.731	&	0.646	\\
            41	&	\textbf{Cornell-BioLay} 	&	0.601	&	0.333	&	0.574	&	0.630	&&	0.980	&	0.508	&	0.797	&	0.891	&&	0.349	&	0.156	\\
            42	&	\textbf{xpc}	&	0.821	&	0.666	&	0.770	&	0.686	&&	0.884	&	0.372	&	0.411	&	0.831	&&	0.285	&	0.091	\\
            43	&	\textbf{Hemlo}	&	0.190	&	0.245	&	0.210	&	0.076	&&	0.766	&	0.857	&	0.408	&	0.258	&&	0.929	&	0.652	\\
            44	&	\textbf{anjaneya}	&	0.140	&	0.363	&	0.126	&	0.493	&&	0.874	&	0.176	&	0.444	&	0.447	&&	0.637	&	0.745	\\
            45	&	\textbf{Runtime\_Terror}	&	0.630	&	0.524	&	0.639	&	0.499	&&	0.864	&	0.711	&	0.737	&	0.604	&&	0.101	&	0.126	\\
            46	&	\textbf{Abhi\_Sidd} 	&	0.420	&	0.439	&	0.386	&	0.400	&&	0.735	&	0.005	&	0.568	&	0.201	&&	0.724	&	0.401	\\
            47	&	\textbf{cbdch} 	&	0.479	&	0.451	&	0.439	&	0.726	&&	0.846	&	0.224	&	0.399	&	0.911	&&	0.413	&	0.000	\\
            48	&	\textbf{aLoneLM} 	&	0.435	&	0.233	&	0.453	&	0.277	&&	0.976	&	0.691	&	\textbf{1.000}	&	0.520	&&	0.129	&	0.175	\\
            49	&	\textbf{hohoho}	&	0.339	&	0.124	&	0.365	&	0.212	&&	0.971	&	0.642	&	0.874	&	0.647	&&	0.045	&	0.139	\\
            50	&	\textbf{huizige}	&	0.499	&	0.410	&	0.466	&	0.395	&&	0.809	&	0.187	&	0.277	&	0.578	&&	0.350	&	0.104	\\
            51	&	\textbf{SSS} 	&	0.000	&	0.187	&	0.000	&	0.356	&&	0.874	&	0.000	&	0.299	&	0.532	&&	0.598	&	0.371	\\
            52	&	\textbf{H2P}	&	0.013	&	0.000	&	0.025	&	0.174	&&	0.765	&	0.089	&	0.000	&	0.694	&&	0.269	&	0.053	\\
            53	&	\textbf{KnowLab}	&	0.283	&	0.283	&	0.230	&	0.000	&&	0.000	&	0.158	&	0.971	&	0.000	&&	0.000	&	0.222	\\
    \end{tabular}
    }
    \caption{Task leaderboard with min-max normalization. \textbf{R} = ROUGE F1, \textbf{BertS} = BertScore, \textbf{FKGL} = Flesch-Kincaid Grade Level, \textbf{DCRS} = Dale-Chall Readability Score, \textbf{CLI} = Coleman-Liau Index, \textbf{AlignS} = AlignScore.
    }
    \label{tab:st1_results_normalized}
\end{table*}

\section{Task Results} \label{sec:results}

    
Table \ref{tab:st1_results} presents the performance of the submission selected by each team to appear on the final leaderboard, according to the defined task metrics.  


Overall, the final leaderboard of BioLaySumm 2024 contains a total of 53 teams, who made a combined total of over 200 submissions. This represents a 165\%  increase in participation over BioLaySumm 2023, which attracted a total of 20 teams across two subtasks. 
In this section, we summarize some of the key results and notable trends that were observed among participants.

\paragraph{Model selection trends}
We identify several trends amongst participants in terms of the models used for experiments.\footnote{Information about model selection is derived from both system papers submitted by participants and a ``system description" field included in the submission form on CodaBench.} Firstly, the use of Large Language Models was found to be particularly prevalent, with a total of 18 teams indicating that LLMs were used in some capacity. Compared to the 3 teams who used LLMs in BioLaySumm 2023, this represents a stark increase that is reflective of shifts in the broader research landscape of NLP. Within those teams using LLMs, biomedical-specific models such as BioGPT \citep{Luo2022BioGPTGP} and BioMistral \citep{labrak2024biomistral} proved popular, with 7 teams indicating they used such models. Other LLMs used include GPT-4 (2), LLAMA (2), and Claude (1). There is evidence that LLMs were used for both summary generation and summary post-processing, with various settings (including fine-tuned, few-shot, and zero-shot) being adopted.

Outside of LLMs, the T5 \citep{Raffel2019ExploringTL} model family proved the most popular alternative approach, with 13 teams making use of these models in their selected submissions. In particular, the FLAN-T5 \cite{Chung2022ScalingIL} model was found to be widely-used, being selected by 9 teams. Interestingly, only 3 teams were found to have used BART-based models, a significant drop from the previous BioLaySumm edition, where they were the most widely adopted approach. We find this shift in model selection to be an encouraging sign that participants are keen to explore novel methods for Lay Summarisation, in line with our overall task objectives.       

\paragraph{Baseline comparison}
As shown in Table \ref{tab:st1_results}, 5 teams exceed the overall rank of the BART baseline system. This represents an an increase on the previous edition of the task, whereby only 1 team outperformed the same baseline system in terms of overall ranking.


\paragraph{Number of models used} 
Contrary to the previous task edition, we find that more teams opted for the use of a single unified model for both datasets (27 out of 53), as opposed to using one model for each dataset. 
This is likely a result of a significant increase in the use of Large Language Models, an unsurprising shift that reflects the current research landscape in Natural Language Processing.
Interestingly, the top four ranked teams can all be seen to adopt a 2-model approach, indicative of the potential benefits of having a distinct model specifically catering to the different lay summary styles of each dataset.

\paragraph{Use of additional data}
As with this previous task edition, we found that very few teams opted to make use of additional data (i.e., data not provided by the organizers as part of the task) in model development. As shown by the \textbf{+} column in Table \ref{tab:st1_results}, only three teams - \textbf{AUTH}, \textbf{elirf}, and \textbf{KnowLab} - indicated that they adopted such an approach.




\paragraph{Reflection on evaluation protocol changes}
Here, we discuss the impact of the changes made to the evaluation protocol over the previous task edition. As mentioned in \S\ref{sec:eval}, the first of these changes surrounds the introduction of new metrics for the Readability and Factuality criteria. 
As a model-based simplification metric, LENS was introduced to provide an additional angle for teams to consider for Readability, with \citet{maddela-etal-2023-lens} demonstrating that the metric correlates particularly well with the \textit{fluency} ratings of human annotators for simplified texts. Notably, LENS does not exhibit a strong alignment with other (more heuristic) Readability metrics, suggesting that these metrics may not capture this aspect of simplified texts. 

For Factuality, we introduced the AlignScore and SummaC metrics as a replacement for a fine-tuned version of BARTScore to avoid potential bias toward BART-based models. However, given that these metrics broadly involve comparing a generated summary to the source text, these metrics tend to favor highly extractive outputs. Given that reference lay summaries tend to be quite abstractive (particularly in the case of the eLife dataset), this resulted in a trade-off between scoring highly for Factuality and the metrics of Relevance or Readability. Overall, we observe that the systems that ranked the highest were those that most successfully balanced this trade-off, typically obtaining strong Relevance and Readability scores while maintaining relatively high Factuality scores.

Finally, the process for the calculation of final rankings was changed from summing individual criterion rankings to the averaging of average criterion scores. This change was motivated by the failure of the previous method of ranking to take into account the relative difference between average scores for a given criterion, something that was commented on by last year's participants.\footnote{For example, in terms of average criterion score, the team ranked 1st may outperform the team ranked 2nd by a large margin, who in term may outperform the team ranked 3rd by a small margin. However, by converting these scores to rankings, all differences are treated as equal.} However, the new ranking system was also found to be not without its issues, particularly surrounding the existence of outliers. Specifically, it was observed that, if there existed teams that scored particularly poorly for a given metric, then all other teams would obtain relatively strong (and less diverse) scores for this metric relative to others - this can be seen for the FKGL metric in Table \ref{tab:st1_results_normalized}.

\section{Submissions} \label{sec:submission}
Out of the 53 participating teams, 14 teams submitted system papers. Here, we provide a brief summary of the approaches taken by these teams.

\paragraph{UIUC\_BioNLP} \citep{UIUC} This team produced the top-ranked submission, adopting an extract-then-summarize approach that utilizes TextRank \citep{mihalcea-tarau-2004-textrank} for salient sentence extraction, followed by a fine-tuned GPT-3.5-turbo model for summary generation. Specifically, their submitted system extracted the top 40 most salient sentences using TextRank, and their GPT-bsaed model is fine-tuned on 200 examples.
Additional experimentation was conducted using various extractive summarization approaches and comparing the number of examples required for effective fine-tuning. Furthermore, the team also explored a LongFormer-based approach that further incorporates retrieved Wikipedia data in a Retrieval-Augmented Generation (RAG) setup.

\paragraph{Cytun AI} \citep{Ctyun} Making the second-ranked submission, the methodology of this team surrounds the use of fine-tuned LLMs. As part of their experimentation, they compare two approaches for handling lengthy input articles: hard truncation and text chunking. Additionally, their summary-generation pipeline includes data preprocessing, augmentation, and prompt engineering.    
\paragraph{Saama Technologies} \citep{Saama} This team achieved the third-ranking submission, which surrounded fine-tuning a Mistral-7B model\footnote{\url{mistral-7B-instruct-v0.2 }} in an unsupervised fashion using low-ranked adaptation (LoRA) \citep{Hu2021LoRALA}, followed by zero-shot summary generation and post-processing to remove redundant sentences. This team also experiments with several other fine-tuning methods, including supervised fine-tuning with LoRA and Direct Preference Optimization \citep{rafailov2023direct}.

\paragraph{WisPerMed} \citep{WisPerMed} Ranking in fourth place, the selected submission of WisPerMed utilized a fine-tuned BioMistral model, combined with few-shot prompting and a Dynamic-Expert selection (DES) mechanism. Specifically, their BioMistral Model was trained using abstracts and lay summaries of the provided train set; and their proposed DES mechanism involved generating several lay summary versions with different prompts for a given input, before selecting the most desirable based on the scores of the references-less Readability and Factuality metrics used in the task.
In additional experiments, they also measured system performance utilizing LLAMA3, as well as that of few-shot and zero-shot model variants. 
The task organizers selected this team to receive an award for the ``most innovative approach".

\paragraph{AUTH} \citep{AUTH} Being one of the only teams to utilize external data, this retrieves 300 abstract-lay abstract pairs scraped from the Science Journal for Kids website.\footnote{\url{https://sciencejournalforkids.org/}} They use this retrieved data as in-context examples for GPT-4, which they prompt to augment the provided datasets by rewriting reference summaries with higher readability scores. Finally, they use this data to fine-tune to fine-tune BioBART \citep{yuan-etal-2022-biobart}, whilst also experimenting with controllable generation techniques in the form of control tokens prepended to the input article (\texttt{<elife>} / \texttt{<plos>} and \texttt{<general\_lay\_summary>} /  \texttt{<kids\_lay\_summary>}).
 
\paragraph{Eulerian} \citep{Eulerian} The team experimented with different combinations of the FLAN-T5 \citep{chung2022scaling} model variations and data selection. They compare the performance of these methods with a preposed ``Preprocessing over Abstract" technique, in which they use a regular expression to remove some abstract information (i.e., anything inside of parentheses, braces and brackets), finding that this outperforms all neural methods tested in terms of Relevance and Factuality metrics. 

\paragraph{BioLay\_AK\_SS} \citep{BioLay_AK_SS} Focusing largely on data augmentation, this team generated additional summary samples using 2 general-purpose models: BART \citep{lewis-etal-2020-bart} and PEGASUS \citep{pegasus2020}. The augmented dataset was then used to fine-tune a domain-specific BioBART model, which was found to improve its overall improved overall performance. 

\paragraph{HULAT-UC3M} \citep{HULAT-UC3M} Again comparing the performance of domains-specific and general-purpose models, this team experimented with fine-tuning both Longformer \citep{Beltagy2020Longformer} and BioBART models on the given datasets. Additionally, they experiment with extending BioBART to utilise Longformer-based sparse attention, thus allowing it to process longer inputs. Overall, they found that fine-tuning the standard BioBART model on each dataset yields the best performance.

\paragraph{DeakinNLP} \citep{DeakinNLP} This team assessed the performance of both a fine-tuned Longformer and GPT-4 (with zero- and few-shot prompting). Additional analysis is also conducted surrounding data selection and the performance vs. cost trade-off between select methods.

\paragraph{elirf} \citep{ELiRF} Again utilising Longformer as their base model, this team experimented with domain-adaption via a continuous pre-training approach. During pre-training, several pretraining tasks were aggregated to inject linguistic knowledge and increase the abstractiveness of generated summaries. Finally, they developed a regression-based ranking model that improved system performance by selecting the most promising from a set of generated summaries.

\paragraph{RAG-RLRC-LaySum} \citep{RAG-RLRC-LaySum} This team developed a Retrieval-Augmented Generation (RAG) Lay Summarisation approach, utilizing multiple knowledge sources (including both source documents and Wikipedia). They experiment with LLMs (Gemini and ChatGPT) for both summary generation and paraphrasing, in addition to a Longformer baseline.
Lastly, the team also develop a Reinforcement Learning strategy to fine-tune the readability of generated summaries.   

\paragraph{SINAI} \citep{SINAI} Focusing largely on a few-shot setting, this team compared the performance of several popular LLMs including GPT-3.5, GPT-4, and LLAMA3. Further experimentation surrounded the fine-tuning of a smaller LLAMA model (LLAMA3-8B) using both parameter-efficient LoRA techniques and Direct Preference Optimization \citep{rafailov2023direct}.

\paragraph{XYZ} \citep{XYZ} This team performed a thorough comparison of several state-of-the-art LLMs, focusing largely on comparing the readability of generated summaries. Further experimentation surrounds Summary rewriting, Title infusing, K-shot prompting, and LoRA-based fine-tuning, with their best-performing submission utilizing a combination of these methods and obtaining the best overall Readability scores.

\paragraph{HGP\_NLP} \citep{HGP-NLP} This team fine-tune and evaluate multiple T5 model variants, also experimenting with LoRA-based fine-tuning.

\section{Conclusion}

The second edition of the BioLaySumm Shared Task was hosted by the BioNLP Workshop@ACL 2024. Several changes were implemented over the previous edition of the task covering participation rules, evaluation metrics, and ranking protocol. 
In terms of participant engagement, the task attracted a total of 53 teams, representing a significant growth from the previous edition's 20 teams. Our results indicate a drastic shift towards the use of LLMs for lay summarisation, with a wide range of both domain-specific and general-purpose LLMs being adopted in various settings across participant submissions.  



\bibliography{anthology,custom}
\bibliographystyle{acl_natbib}

\appendix
\section{Appendix}

\begin{table}[]
    \centering
    \resizebox{\columnwidth}{!}{
    \begin{tabular}{rlcccc}
        \hline \textbf{$\star$} & \textbf{Team} & \textbf{Relevance} & \textbf{Readability} & \textbf{Factuality} & \textbf{Avg.} \\
           \hline
            1	&	\textbf{UIUC\_BioNLP}	&	0.998	&	0.712	&	0.686	&	0.799	\\
            2	&	\textbf{Ctyun AI}  \	&	0.976	&	0.661	&	0.728	&	0.788	\\
            3	&	\textbf{Saama Technologies} 	&	0.977	&	0.730	&	0.657	&	0.788	\\
            4	&	\textbf{WisPerMed} 	&	0.948	&	0.750	&	0.655	&	0.784	\\
            5	&	\textbf{cylaun} 	&	0.902	&	0.713	&	0.694	&	0.770	\\
            6	&	\underline{\textbf{BART Baseline}}	&	0.901	&	0.655	&	0.631	&	0.729	\\
            7	&	\textbf{AUTH}	&	0.922	&	0.713	&	0.552	&	0.729	\\
            8	&	\textbf{maverick}	&	0.737	&	0.519	&	0.911	&	0.723	\\
            9	&	\textbf{Empress}	&	0.753	&	0.772	&	0.563	&	0.696	\\
            10	&	\textbf{eulerian}	&	0.637	&	0.507	&	0.933	&	0.692	\\
            11	&	\textbf{BioLay\_AK\_SS}	&	0.732	&	0.545	&	0.793	&	0.690	\\
            12	&	\textbf{HULAT-UC3M}	&	0.952	&	0.622	&	0.485	&	0.686	\\
            13	&	 \textbf{Atif\_Tanish}	&	0.738	&	0.772	&	0.547	&	0.686	\\
            14	&	\textbf{qwerty}	&	0.509	&	0.631	&	0.891	&	0.677	\\
            15	&	\textbf{Deakin}	&	0.912	&	0.592	&	0.506	&	0.670	\\
            16	&	\textbf{MDSCL}	&	0.743	&	0.579	&	0.681	&	0.667	\\
            17	&	\textbf{MDS-CL}	&	0.729	&	0.570	&	0.703	&	0.667	\\
            18	&	\textbf{elirf}	&	0.890	&	0.572	&	0.529	&	0.664	\\
            19	&	\textbf{RAG-RLRC-LaySum}	&	0.821	&	0.643	&	0.522	&	0.662	\\
            20	&	\textbf{naive\_bhais}	&	0.790	&	0.565	&	0.626	&	0.660	\\
            21	&	\textbf{MDS-CL}	&	0.714	&	0.788	&	0.418	&	0.640	\\
            22	&	\textbf{MDS-CL} 	&	0.753	&	0.734	&	0.401	&	0.629	\\
            23	&	\textbf{DhruvShlo}	&	0.666	&	0.723	&	0.497	&	0.628	\\
            24	&	\textbf{naman\_tejas}	&	0.600	&	0.699	&	0.585	&	0.628	\\
            25	&	\textbf{SINAI}	&	0.743	&	0.746	&	0.386	&	0.625	\\
            26	&	\textbf{XYZ}	&	0.658	&	0.819	&	0.384	&	0.620	\\
            27	&	\textbf{gpsigh}	&	0.379	&	0.585	&	0.893	&	0.619	\\
            28	&	\textbf{YXZ}	&	0.685	&	0.840	&	0.319	&	0.614	\\
            29	&	\textbf{sanika}	&	0.617	&	0.262	&	0.922	&	0.600	\\
            30	&	\textbf{Bossy Beaver} 	&	0.662	&	0.609	&	0.528	&	0.599	\\
            31	&	\textbf{Dayal K-Laksh G}	&	0.466	&	0.409	&	0.895	&	0.590	\\
            32	&	\textbf{MKGS}	&	0.492	&	0.333	&	0.926	&	0.584	\\
            33	&	\textbf{Shallow-Learning}	&	0.652	&	0.769	&	0.328	&	0.583	\\
            34	&	\textbf{NLPSucks} 	&	0.387	&	0.830	&	0.530	&	0.582	\\
            35	&	\textbf{CookieMonster}	&	0.671	&	0.704	&	0.369	&	0.581	\\
            36	&	\textbf{NoblesseUranium}	&	0.588	&	0.572	&	0.577	&	0.579	\\
            37	&	\textbf{roon}	&	0.727	&	0.822	&	0.166	&	0.572	\\
            38	&	\textbf{jimmyapples} 	&	0.701	&	0.827	&	0.187	&	0.572	\\
            39	&	\textbf{Shivam} 	&	0.398	&	0.417	&	0.884	&	0.566	\\
            40	&	\textbf{HGP\_NLP}	&	0.311	&	0.684	&	0.688	&	0.561	\\
            41	&	\textbf{Cornell-BioLay} 	&	0.535	&	0.794	&	0.253	&	0.527	\\
            42	&	\textbf{xpc}	&	0.736	&	0.625	&	0.188	&	0.516	\\
            43	&	\textbf{Hemlo}	&	0.180	&	0.572	&	0.791	&	0.514	\\
            44	&	\textbf{anjaneya}	&	0.281	&	0.485	&	0.691	&	0.486	\\
            45	&	\textbf{Runtime\_Terror}	&	0.573	&	0.729	&	0.113	&	0.472	\\
            46	&	\textbf{Abhi\_Sidd} 	&	0.411	&	0.378	&	0.563	&	0.450	\\
            47	&	\textbf{cbdch} 	&	0.524	&	0.595	&	0.207	&	0.442	\\
            48	&	\textbf{aLoneLM} 	&	0.349	&	0.797	&	0.152	&	0.433	\\
            49	&	\textbf{hohoho}	&	0.260	&	0.783	&	0.092	&	0.378	\\
            50	&	\textbf{huizige}	&	0.443	&	0.463	&	0.227	&	0.378	\\
            51	&	\textbf{SSS} 	&	0.136	&	0.426	&	0.484	&	0.349	\\
            52	&	\textbf{H2P}	&	0.053	&	0.387	&	0.161	&	0.200	\\
            53	&	\textbf{KnowLab}	&	0.199	&	0.282	&	0.111	&	0.197	\\
    \end{tabular}
    }
    \caption{Task leaderboard with min-max normalization. The $\star$ column denotes the submission rank. \textbf{R} = ROUGE F1, \textbf{BertS} = BertScore, \textbf{FKGL} = Flesch-Kincaid Grade Level, \textbf{DCRS} = Dale-Chall Readability Score, \textbf{CLI} = Coleman-Liau Index, \textbf{AlignS} = AlignScore.
    }
    \label{tab:st1_results_averages}
\end{table}

\end{document}